\documentclass[letterpaper, 10 pt, conference]{ieeeconf}  

\IEEEoverridecommandlockouts                              

\usepackage{graphicx}
\usepackage{float}
\usepackage{amsmath}
\usepackage[section]{placeins}
\usepackage{lipsum} 
\usepackage{environ} 
\usepackage{amsmath}
\usepackage{tabularx}
\usepackage{adjustbox}
\usepackage{comment}
\usepackage{hyperref}
\usepackage{algorithmic}
\usepackage{algorithm}

\newcommand*{\tabbox}[2][t]{%
    \vspace{-150pt}\parbox[#1][3.7\baselineskip]{1cm}{\strut#2\strut}}

\title{\LARGE \bf
A Benchmark Comparison of Visual Place Recognition Techniques for Resource-Constrained Embedded Platforms
}

\author{Rose Power$^{1}$, Mubariz Zaffar$^{2}$, Bruno Ferrarini$^{1}$, Michael Milford$^{3}$, Klaus McDonald-Maier$^{1}$ and Shoaib Ehsan$^{1}$
\thanks{*This work was supported by the UK Engineering and Physical Sciences Research Council through Grants EP/R02572X/1, EP/P017487/1 and in part by the RICE project funded by the National Centre for Nuclear Robotics Flexible Partnership
Fund. \textit{(Corresponding author: Shoaib Ehsan)}}
\thanks{$^{1}$R. Power, B. Ferrarini, K. McDonald-Maier and S. Ehsan are with the School of Computer Science and Electronic Engineering, University of Essex, Colchester CO4 3SQ, United Kingdom
        {\tt\small (e-mail: rp17188@essex.ac.uk; bferra@essex.ac.uk; kdm@essex.ac.uk; sehsan@essex.ac.uk)}}%
\thanks{$^{2}$M. Zaffar is with the Delft University of Technology, 2628 CD Delft, Netherlands
        {\tt\small (e-mail: m.zaffar@tudelft.nl)}}%
\thanks{$^{3}$M. Milford is with the School of Electrical Engineering and Computer
Science, Queensland University of Technology, Brisbane, QLD 4000, Australia
        {\tt\small (e-mail: michael.milford@qut.edu.au)}}%
}

\begin{document}

\maketitle
\thispagestyle{empty}
\pagestyle{empty}

\begin{abstract}
Visual Place Recognition (VPR) has been a subject of significant research over the last 15 to 20 years. VPR is a fundamental task for autonomous navigation as it enables self-localization within an environment. Although robots are often equipped with resource-constrained hardware, the computational requirements of and effects on VPR techniques have received little attention. In this work, we present a hardware-focused benchmark evaluation of a number of state-of-the-art VPR techniques on public datasets. We consider popular single board computers, including ODroid, UP and Raspberry Pi 3, in addition to a commodity desktop and laptop for reference. We present our analysis based on several key metrics, including place-matching accuracy, image encoding time, descriptor matching time and memory needs.
Key questions addressed include: (1) How does the performance accuracy of a VPR technique change with processor architecture? (2) How does power consumption vary for different VPR techniques and embedded platforms? (3) How much does descriptor size matter in comparison to today's embedded platforms' storage? (4) How does the performance of a high-end platform relate to an on-board low-end embedded platform for VPR? The extensive analysis and results in this work serve not only as a benchmark for the VPR community, but also provide useful insights for real-world adoption of VPR applications.

\end{abstract}


\section{Introduction} \label{introduction}
Recalling a previously visited place using only visual information has become a subject of interest within the robotic vision community and therefore Visual Place Recognition (VPR) has developed as a dedicated field within autonomous robotics over the past 15 years \cite{lowry2015visual}.  VPR is a key part of many processes, including Simultaneous Localisation and Mapping (SLAM) \cite{cadena2016past}. VPR is a challenging task to perform accurately and consistently. A VPR system must be able to correctly identify whether, given an image, the location has been visited before, and be able to correctly identify and handle false positive results. In addition, to be useful in autonomous navigation, a VPR system must be able to run in real-time \cite{maffra2019real}. There are many factors that can affect VPR. External factors include various appearance and viewpoint variations. Whereas internal factors are usually due to architecture or platform, and as such can include factors such as computational power and memory capacity.

Over the years, many techniques have been developed to tackle the deficiencies of VPR. Many of these techniques were designed to address one of the individual or a combination of problems. It is often useful to compare these VPR techniques to discern which technique is most suited to a specific task or situation. Several works compare different VPR techniques \cite{zaffar2019levelling} \cite{zaffar2019state}. Hulens et al. works to produce an overview of the best CPU-based processing platforms for complex image processing on-board a UAV \cite{hulens2015choose}. However, the subject of implementing and comparing VPR on resource constrained embedded platforms, which is potentially very useful for the application on VPR onboard small unmanned aerial vehicles, is relatively unexplored.
This research compares the performance of several state-of-the-art VPR techniques when implemented on a number of popular embedded platforms. The effects that a resource-constrained platform can have on a variety of VPR techniques will be explored, including the clear variation discovered in board architectures and difference in CPU utilization between embedded and non-embedded platforms, which will be discussed later in this paper.

The remainder of the paper is organized as follows. In Section \ref{literaturereview}, a comprehensive literature review regarding the VPR state-of-the-art and the existing evaluation-based works is presented. Section \ref{experimentalsetup} presents the details of the experimental setup designed for the evaluation performed in this work. Section \ref{resultsandanalysis_new} puts forth the results and analysis obtained by evaluating contemporary VPR techniques on public VPR datasets for a number of hardware platforms. Finally, conclusions and future directions are presented in Section \ref{conclusionsandfuturework}.  
  
\section{Literature Review} \label{literaturereview}
\subsection{VPR Techniques}
VPR generally works by extracting feature descriptors of a reference image and a query image, either local or global depending on method type, and comparing these descriptors to determine whether a match between reference and query image is successful. The method of the comparison of feature descriptors is determined by the VPR technique in use.
Some of the early techniques employed for VPR made use of handcrafted high-quality feature descriptors such as SURF \cite{bay2006surf} and SIFT \cite{lowe2004distinctive}. These handcrafted feature descriptors are classified as either local or global feature descriptors. SIFT \cite{lowe2004distinctive} makes use of key points extracted from an image using difference-of-gaussians. SIFT features were used for VPR in \cite{stumm2013probabilistic}. SURF \cite{bay2006surf} is a modified version of SIFT. It makes use of Hessian-based detectors as opposed to Harris detectors like traditional SIFT. SURF features were used for VPR in \cite{murillo2007surf}. SIFT and SURF are examples of local image descriptors. Contrary, Gist \cite{oliva2006building} is a global feature detector that employs Gabor filters to summarise gradient information from an image. Gist was used for image matching in \cite{murillo2009experiments} and \cite{singh2010visual}.

Histogram-of-orientated-gradients (HOG) \cite{dalal2005histograms} \cite{freeman1995orientation} is a computer vision descriptor that makes use of gradients that are stored in a histogram and is often used as global descriptors. HOG was used for VPR in \cite{mcmanus2014scene}. More recently, CoHOG \cite{zaffar2020cohog} was designed to make use of HOG \cite{dalal2005histograms} \cite{freeman1995orientation} descriptors to represent regions of interest. CoHOG was designed to achieve lateral viewpoint tolerance. 

Much like other applications \cite{tolias2015particular} \cite{liu2016cross}, the use of convolutional neural networks (CNNs), convolution auto-encoders (CAEs) and deep/shallow neural nets has demonstrated superior results for VPR than those achieved by handcrafted feature-based approaches. The use of neural networks was studied in \cite{chen2014convolutional} where a pre-trained CNN was used to extract features from layers of an input image. These features were then used for image comparison in \cite{sermanet2014overfeat}. Following this the authors of \cite{chen2014convolutional} trained two CNNs for VPR on the specific places dataset (SPED) in \cite{chen2017deep}. Both of theses CNNs, AMOSNet and HybridNet, achieved state-of-the-art performance. In addition, both CNNs have the same architecture as CaffeNet \cite{krizhevsky2012imagenet}. However, where the weights used for AMOSNet are randomised, the weights used for HybridNet are taken from CaffeNet trained on the ImageNet dataset \cite{deng2009imagenet}.

The design of CNN layer activations is also something that has been extensively studied, specifically pooling approaches that are employed on convolution layers. Such pooling approaches include; Max-Pooling \cite{tolias2015particular}, Cross-Pooling \cite{liu2016cross}, Sum-Pooling \cite{babenko2015aggregating} and Spatial Max-Pooling \cite{jaderberg2015spatial}. 

A Vectors of Locally Aggregated Descriptors (VLAD) layer was added to CNN architecture in \cite{arandjelovic2016netvlad}, to allow for end-to-end training specifically for use in VPR. This new VLAD layer was implemented in several different CNN models, including AlexNet \cite{krizhevsky2012imagenet} and VGG-16 \cite{simonyan2014very}.

It has been suggested using regions-based descriptions of images could increase the matching performance of VPR \cite{chen2018learning} \cite{khaliq2018holistic} \cite{facil2019condition} \cite{hausler2019multi}. A CNN-based regional approach with a VLAD layer, RegionVLAD, was implemented in \cite{khaliq2018holistic}. 

\subsection{Benchmarking}
Benchmark efforts have a significant impact on the understanding of a model’s usability within the target application. In terms of resource-constrained embedded platforms the most prominent benchmarks are power consumption, memory utilization and CPU utilisation. 

Canziani et al. \cite{canziani2016analysis} conducted an analysis of several deep neural network (DNN) architectures. They recorded accuracy, power consumption and memory footprint, as well as the number of parameters and operations count for computer vision tasks. The experimentation was carried out on an NVIDIA Jetson TX1 board. Huang et al. take a similar approach in their exploration of the speed/accuracy trade-off of full image classification in \cite{huang2017speed}. Additionally, they compare performance of an Intel Xeon CPU and a NVIDIA Titan X GPU.

TANGO \cite{karki2019tango} employs the metrics of inference time, power consumption and memory usage to access CNN models implemented on a variety of hardware platforms including an embedded GPU and a FPGA. Palit et al. \cite{palit2019uniform} highlight the importance of energy usage by presenting an energy estimation model and empirical data from several CNNs.

The authors in \cite{fan2007power} show that CPU utilization of running processes is directly related to the power consumption of a CPU. Over extended periods of time, this power consumption becomes a significant factor, especially in battery powered UAVs. This work is extended upon by Zaffar et al. \cite{zaffar2019state}, who investigate the accuracy, processing power consumption and projected memory requirements of several state-of-the-art VPR techniques. The range evaluated includes both CNN-based and feature-based techniques. 

While accuracy, power consumption and memory requirements as studied in this paper have been looked into before, the effect of platform architecture on these measurements when performing VPR has not. In addition, little has been done in benchmarking the potential real-time usefulness of different VPR techniques, as such we work to present a benchmark for this with the introduction of Real-Time Matches Frames (RMF).

\section{Experimental Setup} \label{experimentalsetup}
This section details the computational platforms evaluated, the datasets that are used for the evaluation and the metrics used for evaluation.

\begin{table*}[t!]
\centering
\begin{center}
\caption{Table showing specifications for each board.}
\begin{tabular}{ |c|c|c|c| } 
 \hline
 Name & Processor & Architecture & Memory\\
 \hline
 UP & Intel Atom x5-Z8350 Quad-Core Processor 1.92GHz & x86\verb|_|64 & 4GB\\
 \hline
 Raspberry Pi 3 model B & Quad Core 1.2GHz Broadcom BCM2837 64bit CPU & arm64 & 1GB\\
 \hline
 Odroid XU4 & Samsung Exynos5422 ARM Cortex-A15 Quad 2.0GHz & armhf & 2GB\\ 
 \hline
 Laptop & Intel(R) Core i7-8565U CPU 1.8GHz 1.99GHz & x86\verb|_|64 & 16GB\\
 \hline
 Desktop & AMD RYZEN 1400 Quad-Core Processor 3.20 GHz & x86\verb|_|64 & 32GB\\
 \hline
\end{tabular}
\end{center}
\label{table:Boards}
\end{table*}

\begin{figure*}
\centering
\begin{tabular}{ c c c c c l}
\includegraphics[width=0.15\textwidth]{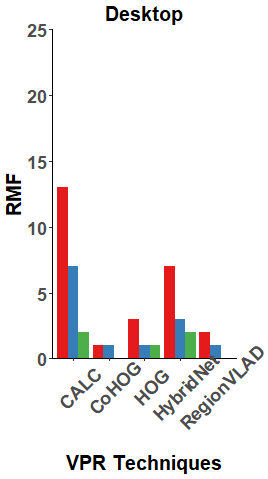} & 
\includegraphics[width=0.15\textwidth]{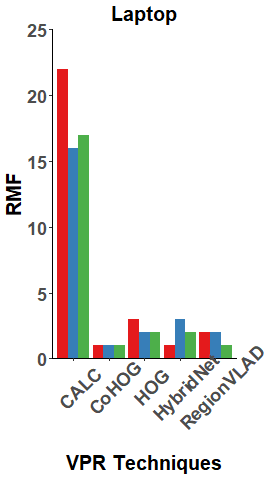}  &
\includegraphics[width=0.15\textwidth]{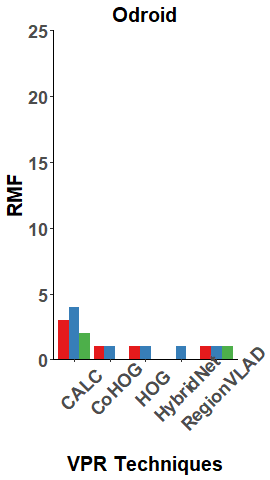} &
\includegraphics[width=0.15\textwidth]{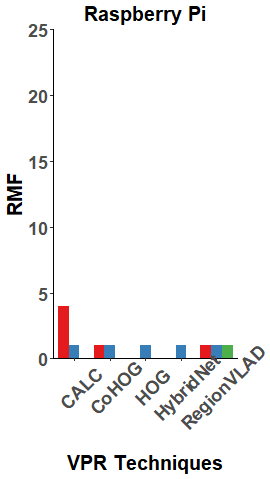} &
\includegraphics[width=0.15\textwidth]{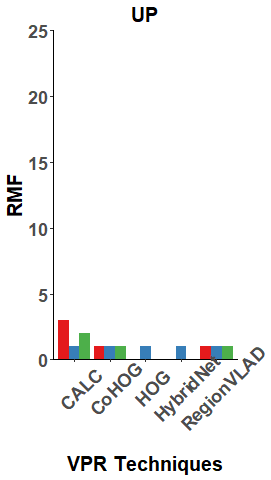} &
\includegraphics[width=0.1\textwidth]{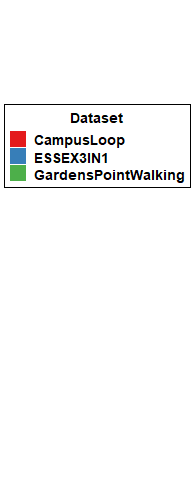}\\ 
\end{tabular}
\caption{Graphs showing the Real-Time Matched Frames (RMF) for every VPR technique used in the paper, on each board. CALC achieved the highest RMF across all platforms, greatly out-performing the other techniques. Of the platforms, the laptop demonstrated the highest number of successfully matched frames within the real-time constraints.}
\label{fig:RMF}
\end{figure*} 

\subsection{VPR Techniques}
The experiments in this paper are carried out on a selection of VPR techniques, as a representation of a variety of technique classes. HOG and CoHOG are training-less handcrafted feature-based approaches. HybridNet and RegionVLAD each use a CNN as an image feature extractor. Additionally, CALC utilizes an auto-encoder to extract image features.
\subsubsection{HOG}
Histogram-of-orientated gradients (HOG) \cite{dalal2005histograms} \cite{freeman1995orientation}  is one of the most widely used VPR techniques that uses hand-crafted feature descriptors. This technique calculates a gradient for every pixel in an image and organises these gradients into the bins of a histogram. These bins will contain the sum of the gradient magnitudes. The technique then uses the cosine function to compare the query and reference images. For our implementation of HOG, we use a cell size of $8\times8$, a block size of $16\times16$, and total of 9 histogram bins as suggested by the authors in \cite{zaffar2019levelling}. 
\subsubsection{CoHOG}
CoHOG \cite{zaffar2020cohog} uses handcrafted feature-based technique that uses image-entropy to extract regions of interest. HOG \cite{dalal2005histograms} \cite{freeman1995orientation} descriptors are then assigned to represent each region and the regional descriptors are compared using cosine matching to achieve lateral viewpoint tolerance. One of the inspirations behind this technique was CNN-based techniques’ ability to extract regions of interest. CoHOG was developed to achieve state-of-the-art performance without any training requirements, unlike traditional CNN-based techniques.
\subsubsection{HybridNet}
HybridNet was trained on the Specific Places Dataset (SPED), and the model weights of the top 5 HybridNet convolutional layers are initialised from CaffeNet trained on the ImageNet dataset. Our implementation employs a spatial pyramidal pooling on activations from conv5 layer to form feature descriptors. L1-difference is then used to match the query and reference images. 
\subsubsection{RegionVLAD}
Region-VLAD \cite{khaliq2018holistic} uses a lightweight CNN-based regional approach, as well as VLAD, to overcome the practical deployment limitations of traditional CNNs used for VPR, due to their computational complexity and significant memory overhead. For our implementation we employ the convolutional layer conv4 of HybridNet, along with 400 region of interests (ROIs). A 256 visual word dictionary is used to extract VLAD descriptors and cosine-similarity is used to match query and reference images.
\subsubsection{CALC}
CALC was first introduced by Merrill et al. when they trained an autoencoder in an unsupervised manner for the first-time for use in VPR. The objective of the auto-encoder was to re-create the HOG descriptors of an image, when given a distorted version the image. For our implementation we use model parameters from the 100K iteration of the auto-encoder on the Places dataset \cite{zhou2017places}.

\subsection{Computational Platforms}
The experiments in this paper are carried out on a selection of platforms, listed in Table \ref{table:Boards}.

\subsubsection{UP-CHT01-A20-0464-A11 \cite{UPBoard}}
The UP board is marketed as a high performance board with low power consumption. It has a quad-core Atom x5-z8350 processor running at up to 1.92GHz. The UP board is the only board used in this paper that has an Intel processor, like the desktop and laptop standards.

\subsubsection{Raspberry Pi 3 Model B \cite{RaspberryPi3}}
The Raspberry Pi used in this paper has a quad-core ARM processor with arm64 architecture with 1GB RAM and 100MB of swap. Swap space was increased to 2GB to satisfy the memory demands of the various VPR techniques used in our experiment. It should be noted however, that increasing the size of the swap greatly increases the time taken for computation.

\subsubsection{ODROID XU4 \cite{OdroidXU4}}
The Odroid board used in our study had 1GB RAM and no swap as default. Similarly to the Raspberry Pi swap space was increased to 2GB.
The Odroid board is the only embedded platform that has a fan. The fan fires periodically causing a surge in power consumption. While it is possible to disconnect the fan, for the sake of not damaging the board during the course of these experiments, the fan was kept attached.

\subsubsection{Laptop}
This platform is a DELL XPS 13 9380. It has an Intel(R) Core i7-8565U CPU, operating at 1.99GHz, and 16GB of RAM. Due to its large size, weight and power requirements, it is not feasible to use as an embedded platform. As such, it is only included as a reference for the other embedded platforms.

\subsubsection{Desktop}
A desktop computer was included as an additional reference platform. This desktop had an AMD RYZEN  1400 Quad-Core Processor, operating at 3.20 GHz, and 32GB of RAM. 

\subsection{Evaluation Datasets}
There have been many datasets created for the purpose of testing VPR. In this subsection we will introduce the datasets used to test the techniques and platforms introduced in subsection III-A and subsection III-B. 
\subsubsection{Garden Point Walking \cite{phdthesis}}
This dataset was created at the Queensland University of Technology and displays both illumination and viewpoint variation, as well as dynamic objects making it a challenging dataset.
\subsubsection{Campus Loop \cite{merrill2018lightweight}}
The campus loop dataset was created from a sequence of indoors and outdoors images of a campus environment. Whilst the images captured indoors do not present great seasonal variation, the image captured outside present significant seasonal and some illumination variation. In addition, the images also contain viewpoint variation and dynamic objects.
\subsubsection{ESSSEX3IN1 \cite{zaffar2018memorablemaps}}
The ESSEX3IN1 dataset was created to provide both viewpoint and appearance variation. The dataset contains images that are confusing for VPR techniques, with challenging dynamic objects and uninformative scenes, causing most state-of-the-art techniques to struggle.

\begin{figure}[t]
\centering
\includegraphics[width=0.8\columnwidth]{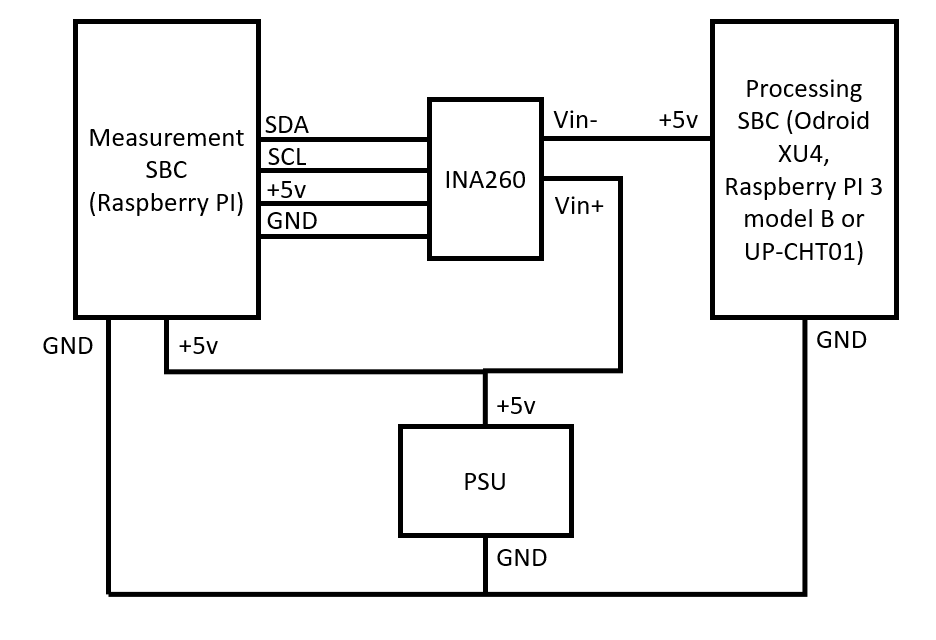}
\caption{Power measurement circuit schematic.}
\label{fig:Circuit}
\end{figure}

\subsection{Evaluation Metrics}
\subsubsection{Matching Performance}
The image with the highest matching score extracted from the mapped reference dataset should correspond to the query image of the same location. As such, a match is determined as correct if, given a query image, the technique assigns the corresponding reference image the highest similarity score, determined for each technique as described in the corresponding experimental setup in section \ref{experimentalsetup}, of every possible image. Determining the percentage of correctly matched images for each of the datasets used will give the accuracy of each VPR technique.

\subsubsection{Algorithm Efficiency}
In this paper we also evaluate the efficiency of each technique used is evaluated by measuring the CPU usage, memory usage and processing time of each technique. CPU usage and memory usage are reported as a percentage. Memory usage is measured as the percentage of active memory out of total memory. In addition, we record processing time, which is reported in seconds. This includes the feature encoding time for an input query image, as well as the descriptor matching time for the number of reference images in the dataset, image loading and preprocessing. In order to benchmark the platforms accurately the images were loaded from a dataset. In real-world applications, the images would be acquired from a camera in real-time.

\subsubsection{Power Consumption}
In addition to evaluating the precision-recall of the techniques produced by each board, the power consumption of each board was measured using a setup of a INA260 module \cite{siepert_wellish_2019} monitored by an additional Raspberry Pi, the confuguration of which is shown in figure \ref{fig:Circuit}. This gave an indication of the power consumption over the course of encoding and matching images. In the configurations tested, the power consumption of each technique was measured whilst encoding and matching each image in the query dataset in turn.

\subsubsection{Real-Time Matched Frames (RMF)}
In a real-time scenario, cameras can provide up to 50 images per second. While this is useful, if the VPR algorithm implemented cannot process images as fast as a camera can provide them, then a decision must be made between processing the next image provided or the image provided at the time. As such an algorithm with high accuracy but also high computational time may be less effective in a real-time scenario than an algorithm with low accuracy and low computational time.

\begin{algorithm}[t]
\small
\caption{Computing RMF}
\label{RMFalg}
\begin{algorithmic}
\STATE Original\verb|_|Matches\verb|_|List = $List_{matches}$
\STATE $N_q = Length(List_{matches})$
\STATE $M_q = Sum(List_{matches})$
\STATE $RMF = 0$
\STATE $V, D, F, K : Given$
\STATE $G : Computed$
\FORALL{$index,$ $element$ $in$ $List_{matches}$}
\IF{$(index + 1)\%G = 0$ \OR $index = 0$}
\IF{$element = 1$}
\STATE{$RMF = RMF + 1$}
\ENDIF
\ENDIF
\ENDFOR
\RETURN $M_q,$ $RMF$
\end{algorithmic}
\end{algorithm}

\begin{figure*}
\centering
\begin{tabular}{ c c c p{0.5cm}}
\includegraphics[width=0.25\textwidth]{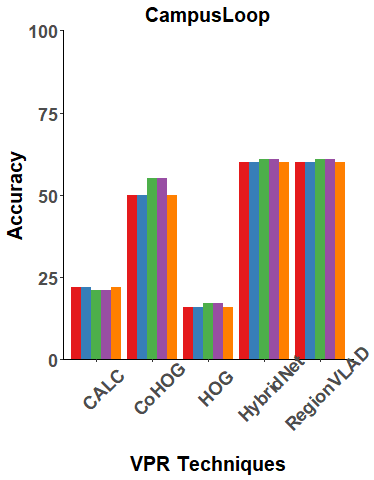} & 
\includegraphics[width=0.25\textwidth]{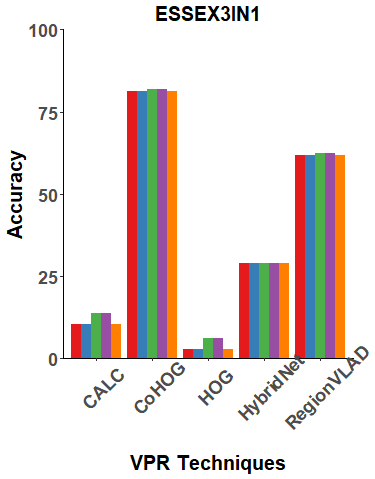}  &
\includegraphics[width=0.25\textwidth]{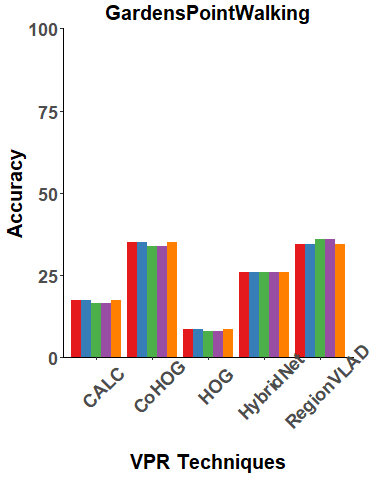} &
\tabbox{\includegraphics[width=0.1\textwidth]{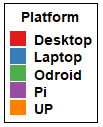}}\\ 
\end{tabular}
\caption{Graphs showing the accuracy for every VPR technique used in the paper, on each board. The results show that across platform the accuracy remains quite consistent. Note that the ARM platforms produced identical results to each other, as did the x86\_64 platforms.}
\label{fig:Accuracy}
\end{figure*}

In this paper we aim to provide a benchmark allowing for the evaluation of algorithms with a view to real-world, real-time VPR. The proposal of our paper is that the critical ratio of incoming frame rate and VPR frame rate will determine if a particular VPR algorithm will perform well in a real-world environment.

Let the frames per second (FPS) sampling rate of the camera be F, which is usually a fixed value. Let the frames sample per distance be D and the platform speed be V. Then the incoming frame rate from the platforms vision sensor can be computed as:
\begin{equation}\label{eq1}
Incoming\ Frame\ Rate = min(K \times D \times V,F)
\end{equation}
Where K is a unit-less constant that represents any further down-sampling cause by the vision pipeline. The VPR frame rate (potential place matched per second) can also be calculated as shown in equation \ref{eq2}, where t\textsubscript{R} is the retrieval time.
\begin{equation}\label{eq2}
VPR\ Frame\ Rate = Floor(\frac{1}{t_{R}})
\end{equation}
The ratio is denoted as G as shown in equation \ref{eq3} and the maximum possible value of G is 1. This is given that the maximum number of VPR query images matched can only must be equal to the total incoming images. Please note that our modelling of Incoming Frame Rate is one traversal-specific. In addition, the Incoming Frame Rate can also be an explicit specification from either the computer vision or a robotics VPR application.
\begin{equation}\label{eq3}
G = Floor(max(\frac{Incoming\ Frame\ Rate}{VPR\ Frame\ Rate}, 1))
\end{equation}
Given the ratio G, a VPR technique will only be able to perform VPR for a query frame after some constant frame interval based on G.

The total correctly matched query images in a VPR dataset at the maximum value of recall can be denoted as M\textsubscript{q} out of a total of N\textsubscript{q} query images in the dataset. Based on the prospective loss of potential place matching candidates for a slow VPR technique due to G $>$ 1, the number of total matched query images will reduce to Real-time Matched Frames (RMF) out of a total of N\textsubscript{q} query images. This RMF can be interpreted as an active evaluation metric and we compute this RMF using Algorithm \ref{RMFalg}.

\section{Results and Analysis} \label{resultsandanalysis_new}
This section contains results based on all evaluation metrics for all computational platforms.

\subsection{Accuracy}
This sub-section reports the findings of the matching performance of each technique on all three datasets used for testing as shown in figure \ref{fig:Accuracy}.

As part of our findings, we discovered that the Raspberry Pi and Odroid produced the same results in terms of accuracy for each VPR technique. Conversely, the UP board, laptop and desktop platforms all produced the same results in terms of accuracy as well. This is likely due to the fact that both the Odroid and Raspberry Pi platforms use ARM processors, having arm64 or armhf architecture, and the UP board, laptop and desktop use Intel Atom, Intel i7 and AMD Ryzen 5 processors respectively, all sharing the x86\verb|_|64 architecture. Ubuntu 18 operating system was used across all platforms. Apart from architecture type, the only difference in OS is that the raspberry Pi uses a 32-bit OS, however it produced they same accuracy as the Odroid board which used a 64-bit OS and had the same ARM architecture suggesting that the difference in 32-bit, 64-bit systems does not have a significant affect on the accuracy of the techniques. The running environments were kept the same for each platform using the same version of Python and additional libraries, with the only difference being when a specific version had to be built for ARM architecture as opposed to the standard x86\verb|_|64 architecture.

\subsection{Algorithm Efficiency}
This sub-section reports the findings of the algorithm efficiency of each platform and technique as performed on all three datasets used for testing.
\subsubsection{CPU Usage}
In terms of CPU usage each platform and technique combination only displayed slight variation between datasets as shown in table \ref{Table:CPU}.

The laptop displayed the highest CPU utilisation by far, using roughly 100\% of the CPU’s capacity for all techniques. In comparison, the Ordoid and Raspberry Pi platforms had very low CPU utilisation for all techniques, using between 15-30\% of their respective CPU’s capacity. In addition, the UP board has relatively low CPU utilisation for all techniques except for CoHOG, which uses an average of 80\% of its CPU’s capacity to perform. The desktop also displayed a high CPU usage when running CoHOG and CALC, but had a relatively low CPU usage for the remaining techniques.

\begin{table*}[htbp]
   \centering
   \caption{CPU usage, memory usage and average processing time of every VPR technique, across all datasets and platforms.}
     \begin{tabular}{l|ccc|ccc|ccc|}
     \multicolumn{1}{r}{} & \multicolumn{3}{c}{CPU} & \multicolumn{3}{c}{Memory} & \multicolumn{3}{c}{Processing Time}  \\
 \cline{2-10} RegionVLAD & \multicolumn{1}{l}{Campus} & \multicolumn{1}{l}{Gardens} & \multicolumn{1}{l|}{ESSEX3IN1} & \multicolumn{1}{l}{Campus} & \multicolumn{1}{l}{Gardens} & \multicolumn{1}{l|}{ESSEX3IN1} &  \multicolumn{1}{l}{Campus} & \multicolumn{1}{l}{Gardens} & \multicolumn{1}{l|}{ESSEX3IN1} \\
    \hline
    \hline
    Desktop  & 34.38     & 35.50    & 27.36     & 66.25     & 71.72     & 79.21     & 1.72    & 1.73    & 2.45  \\
    Laptop  & 100.00     & 99.92    & 99.94     & 69.81     & 69.14     & 81.27     & 1.76     & 1.73     & 2.43 \\
    Odroid  & 14.31     & 17.46    & 18.47     & 53.95     & 51.70     & 51.72     & 7.83     & 8.13     & 8.74 \\
    RPI  & 21.90     & 17.49    & 17.42     & 58.90     & 51.00     & 52.55     & 24.16     & 33.25     & 34.31 \\
    UP  & 27.99     & 28.01    & 26.79     & 70.58     & 82.86     & 78.27     & 10.08     & 10.14     & 16.31  \\
     \hline
     \end{tabular}%
    \\
\begin{tabular}{l|ccc|ccc|ccc|}
     \multicolumn{1}{r}{} & \multicolumn{3}{c}{CPU} & \multicolumn{3}{c}{Memory} & \multicolumn{3}{c}{Processing Time}  \\
 \cline{2-10} CoHOG & \multicolumn{1}{l}{Campus} & \multicolumn{1}{l}{Gardens} & \multicolumn{1}{l|}{ESSEX3IN1} & \multicolumn{1}{l}{Campus} & \multicolumn{1}{l}{Gardens} & \multicolumn{1}{l|}{ESSEX3IN1} &  \multicolumn{1}{l}{Campus} & \multicolumn{1}{l}{Gardens} & \multicolumn{1}{l|}{ESSEX3IN1} \\
    \hline
    \hline
    Desktop  & 91.00     & 96.17    & 89.20     & 64.61     & 70.53     & 78.71     & 0.36    & 0.58    & 0.57  \\
    Laptop  & 100.00     & 100.00    & 100.00     & 68.40     & 67.72     & 82.65     & 0.24     & 0.34     & 0.39 \\
    Odroid  & 16.08     & 20.32    & 20.47     & 45.85     & 46.48     & 53.29     & 8.49     & 18.44     & 17.20 \\
    RPI  & 25.84     & 25.74    & 25.74     & 49.16     & 61.26     & 66.95     & 5.38     & 10.03     & 10.51 \\
    UP  & 79.29     & 86.82    & 79.48     & 68.02     & 80.72     & 78.68     & 1.24     & 1.96     & 2.36  \\
     \hline
     \end{tabular}%
    \\
\begin{tabular}{l|ccc|ccc|ccc|}
     \multicolumn{1}{r}{} & \multicolumn{3}{c}{CPU} & \multicolumn{3}{c}{Memory} & \multicolumn{3}{c}{Processing Time}  \\
 \cline{2-10} CALC & \multicolumn{1}{l}{Campus} & \multicolumn{1}{l}{Gardens} & \multicolumn{1}{l|}{ESSEX3IN1} & \multicolumn{1}{l}{Campus} & \multicolumn{1}{l}{Gardens} & \multicolumn{1}{l|}{ESSEX3IN1} &  \multicolumn{1}{l}{Campus} & \multicolumn{1}{l}{Gardens} & \multicolumn{1}{l|}{ESSEX3IN1} \\
    \hline
    \hline
    Desktop  & 100.14     & 100.14    & 87.07     & 64.95     & 70.34     & 81.67     & 0.07    & 0.07    & 0.12  \\
    Laptop  & 100.00     & 100.00    & 100.00     & 67.87     & 66.64     & 84.84     & 0.17     & 0.14     & 0.21 \\
    Odroid  & 22.70     & 23.27    & 21.35     & 45.15     & 48.31     & 56.47     & 0.68     & 0.69     & 0.70 \\
    RPI  & 28.89     & 29.01    & 29.58     & 55.27     & 66.08     & 70.01     & 0.56     & 0.55     & 0.61 \\
    UP  & 27.57     & 27.71    & 29.30     & 68.05     & 80.57     & 78.40     & 0.67     & 0.68     & 1.04  \\
     \hline
     \end{tabular}%
    \\
\begin{tabular}{l|ccc|ccc|ccc|}
     \multicolumn{1}{r}{} & \multicolumn{3}{c}{CPU} & \multicolumn{3}{c}{Memory} & \multicolumn{3}{c}{Processing Time}  \\
 \cline{2-10} HybridNet & \multicolumn{1}{l}{Campus} & \multicolumn{1}{l}{Gardens} & \multicolumn{1}{l|}{ESSEX3IN1} & \multicolumn{1}{l}{Campus} & \multicolumn{1}{l}{Gardens} & \multicolumn{1}{l|}{ESSEX3IN1} &  \multicolumn{1}{l}{Campus} & \multicolumn{1}{l}{Gardens} & \multicolumn{1}{l|}{ESSEX3IN1} \\
    \hline
    \hline
    Desktop  & 54.10     & 57.08     & 40.07     & 67.69     & 71.16     & 83.02     & 1.18     & 1.17     & 1.85 \\
    Laptop  & 100.00     & 100.00     & 100.00     & 69.06     & 67.86     & 85.55     & 4.81     & 4.80     & 5.54 \\
    Odroid  & 18.83     & 20.80     & 21.36     & 55.35     & 57.17     & 56.30     & 25.56     & 26.21     & 27.21 \\
    RPI  & 25.66     & 25.69     & 25.78     & 68.40     & 68.48    & 71.37     & 20.72     & 20.65     & 20.88 \\
    UP  & 25.95     & 25.97     & 25.57     & 70.94     & 82.85     & 78.49     & 24.57     & 24.47     & 31.38 \\
     \hline
     \end{tabular}%
    \\
\begin{tabular}{l|ccc|ccc|ccc|}
     \multicolumn{1}{r}{} & \multicolumn{3}{c}{CPU} & \multicolumn{3}{c}{Memory} & \multicolumn{3}{c}{Processing Time}  \\
 \cline{2-10} HOG & \multicolumn{1}{l}{Campus} & \multicolumn{1}{l}{Gardens} & \multicolumn{1}{l|}{ESSEX3IN1} & \multicolumn{1}{l}{Campus} & \multicolumn{1}{l}{Gardens} & \multicolumn{1}{l|}{ESSEX3IN1} &  \multicolumn{1}{l}{Campus} & \multicolumn{1}{l}{Gardens} & \multicolumn{1}{l|}{ESSEX3IN1} \\
    \hline
    \hline
    Desktop  & 41.50     & 27.40     & 22.42     & 68.13     & 70.31     & 82.88     & 0.02     & 0.02     & 0.09 \\
    Laptop  & 78.67     & 100.00     & 100.00     & 67.08     & 66.43     & 85.11     & 0.02     & 0.03     & 0.07 \\
    Odroid  & 22.45     & 24.94     & 23.61     & 40.30     & 44.25     & 43.34     & 0.10     & 0.15     & 0.17 \\
    RPI  & 29.36     & 29.00     & 28.98     & 63.38     & 56.41    & 57.96     & 0.20     & 0.27     & 0.30 \\
    UP  & 31.70     & 31.91     & 26.11     & 75.74     & 78.82     & 66.33     & 0.09     & 0.11     & 0.40 \\
     \hline
     \end{tabular}%
    \\
   \label{Table:CPU}%
 \end{table*}%

The embedded platforms demonstrated low CPU utilisation at times. This is likely due to their limited memory and consequent need to page to relatively slow SD card storage. This was not exhibited by the non-embedded platforms, that had enough memory to not require paging. Whilst the need to page caused a slight increase in the computational time, it suggests that potentially a more computationally complex algorithm could be run on the embedded platforms with similar success, given a larger memory capacity.

\subsubsection{Memory Usage}
Trends in memory utilization are difficult to identify due to variation in image resolution between datasets, the amount of memory each board has and the technique used, as shown by the results in table \ref{Table:CPU}. During the experiments in this paper the images were loaded one at a time for processing to cut down on memory consumption given the limited memory capacity of the embedded platforms. As such, no more than one image from the dataset is stored in memory at any given time. Memory utilisation appears also to be affected by platform architecture. 

The ARM architecture platforms utilised the least memory, however this is likely due to the resized and compressed nature of the datasets, required to make the algorithms run on those platforms. It can be speculated that the memory utilisation of the x86 platforms would be similarly low if given the same compressed images. 

\subsubsection{Processing Time}
As expected, the laptop and desktop had a much shorter processing time than any of the embedded platforms. In general, all platforms and techniques took slightly longer when tested with the ESSEX3IN1 dataset, as shown in table \ref{Table:CPU}, likely due to the uninformative images in this dataset. However, the results are similar across all datasets. 

The Raspberry Pi took the longest on all three datasets to complete RegionVLAD, taking over double the time of the next platform. However, in contrast, the Pi had the shortest processing time for HybridNet across all datasets. Despite this the processing time for HybridNet was significantly longer than the other techniques regardless of platform. UP had by far the shortest processing time using CoHOG, greatly out-performing the other embedded platforms in terms of processing time. 

All platforms perform well using both CALC and HOG with the shortest processing times of all the techniques. 

\subsection{Power Consumption}
\begin{figure}[t!]
\centering
\includegraphics[scale=0.4]{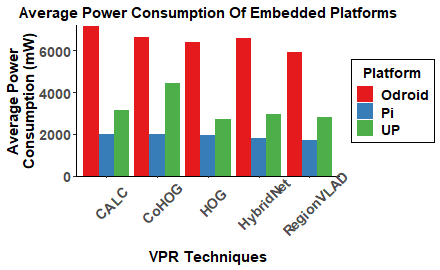}
\caption{Average power consumption of each VPR technique when implemented on the embedded platforms.}
\label{fig:AveragePower}
\end{figure}

This sub-section reports the findings of the power consumption of each technique on all three of the embedded platforms used for testing as shown in figure \ref{fig:AveragePower}.

The Raspberry Pi showed a clear superiority with both low power consumption and relatively low computation time despite being constrained by RAM limitations and having to use swap space on the micro-SD card. This low power consumption is likely due to having an ARM processor and no active cooling unit. 
In comparison the Odroid board consistently demonstrated the highest power consumption per unit time of the three embedded platforms at least partially due to its active cooling unit. This comparatively high-power consumption makes for an unfavourable trade-off between computation time and power consumption. The Odroid board also suffered from RAM limitations, resulting in paging to the micro-SD.
The UP board had a relatively low power consumption per unit time but took the longest on most techniques. Despite taking the longest time for most methods, this is not significant, giving the board an acceptable trade-off between computation time and power consumption. Despite the slightly longer computation time, the UP board was less constrained by RAM than the other embedded platforms.

\subsection{Real-Time Matched Frames (RMF)}
Overall, CALC managed to successful match the most frames across all platforms and datasets, as presented by the graphs in figure \ref{fig:RMF}, often greatly out-performing the other techniques. Out of all platforms, the laptop demonstrated the highest number of successfully matched frames within the real-time constraints, above that of the Desktop described in this paper, despite a supposedly inferior CPU and Memory capacity.
\subsubsection{Campus Loop}
Across all platforms, CALC correctly matched the most frames within the real-time constraints greatly out-performing all the other techniques. RegionVLAD and CoHOG perform relatively consistently across all platforms, albeit quite poorly. The remaining techniques also perform relatively poorly on the Desktop and Laptop platforms however their performance drops when performed on the embedded platforms.
\subsubsection{ESSEX3IN1}
Similarly, to the campus dataset, CALC performs the best on the Desktop, Laptop and Odroid platforms, however on the remaining two platforms CALC has a similar performance as all the other VPR techniques. The other techniques all perform similarly poorly across all platforms.
\subsubsection{GardensPointWalking}
Similar to the other two datasets, CALC again performs the best on all platforms, save for the Raspberry Pi. Aside from RegionVLAD which demonstrates a relatively consistent performance, the remaining techniques struggle greatly on the embedded platforms, unable to successfully match any frames within the real-time constraints.

\section{Conclusions and Future Work} \label{conclusionsandfuturework}
As expected, the overall performance of the high-end platforms was greater than those of the embedded platforms. However, the embedded platforms’ ability to match the accuracy of the high-end platforms indicates good potential for future use in the field of VPR. Out of the three embedded platforms assessed the Raspberry Pi showed clear superiority, with the lowest power consumption, good processing time and similar accuracy to the high-end platforms. Apart from the UP platform, the embedded platforms struggled with descriptor size until additional swap space was added, when performance was acceptable. Descriptor size will become less of a problem as embedded platform’s memory capacity increases.

In terms of power consumption, the Raspberry Pi and Odroid platforms showed the same trends in power consumption per technique, with the most power consumed when using CALC and the least when using RegionVLAD. On the other hand, when the techniques were run on the UP board, CoHOG had the highest power consumption. This is likely due to the UP board utilizing up to 4 cores when performing CoHOG, whereas the Odroid and Pi did not use more than 2 when running the same algorithm. This is potentially due to architectural differences between the x86\verb|_|64 architecture of the UP board and the ARM architecture of the Odroid and Pi.

In terms of RMF it should be noted that a clear limitation is the assumption that the platform performing VPR is moving at a constant velocity. This is not always the case in a real-time real-world scenario. However, for benchmarking purposes, this can be largely mitigated by partitioning the dataset sequence into sub-sequences of near-consistent velocities. In addition, RMF will inherently favour techniques that have their true-positive results distributed throughout the trajectory, as opposed to true-positives concentrated in certain areas, due to distance-based sampling. A better analysis could be obtained by combining RMF with the true-positives (loop-closures) distribution over a dataset trajectory as proposed by Porav et al. \cite{porav2018adversarial}. For applications that require a loop-closure every few meters, as otherwise the localisation error drift becomes too large to handle, the metric proposed in \cite{porav2018adversarial} compliments the value proposed by RMF.

This analysis could be continued as new platforms and platform versions are released, such as the Raspberry Pi 4, with more onboard RAM to help avoid paging to the micro-SD card. Another extension to consider would be a system that selectively applies a high performance VPR algorithm to frames as they come in, combined with a lower performance VPR system that is computationally lightweight and can be applied to every frame.

{
\small
\bibliographystyle{ieeetr}
\bibliography{main}

\begin{thebibliography}{10}

\bibitem{lowry2015visual}
S.~Lowry, N.~S{\"u}nderhauf, P.~Newman, J.~J. Leonard, D.~Cox, P.~Corke, and
  M.~J. Milford, ``Visual place recognition: A survey,'' {\em IEEE Transactions
  on Robotics}, vol.~32, no.~1, pp.~1--19, 2015.

\bibitem{cadena2016past}
C.~Cadena, L.~Carlone, H.~Carrillo, Y.~Latif, D.~Scaramuzza, J.~Neira, I.~Reid,
  and J.~J. Leonard, ``Past, present, and future of simultaneous localization
  and mapping: Toward the robust-perception age,'' {\em IEEE T-RO}, vol.~32,
  no.~6, pp.~1309--1332, 2016.

\bibitem{maffra2019real}
F.~Maffra, L.~Teixeira, Z.~Chen, and M.~Chli, ``Real-time wide-baseline place
  recognition using depth completion,'' {\em IEEE Robotics and Automation
  Letters}, vol.~4, no.~2, pp.~1525--1532, 2019.

\bibitem{zaffar2019levelling}
M.~Zaffar, A.~Khaliq, S.~Ehsan, M.~Milford, and K.~McDonald-Maier, ``Levelling
  the playing field: A comprehensive comparison of visual place recognition
  approaches under changing conditions,'' {\em arXiv preprint arXiv:1903.09107,
  IEEE ICRA Workshop on Database Generation and Benchmarking}, 2019.

\bibitem{zaffar2019state}
M.~Zaffar, A.~Khaliq, S.~Ehsan, M.~Milford, K.~Alexis, and K.~McDonald-Maier,
  ``Are state-of-the-art visual place recognition techniques any good for
  aerial robotics?,'' {\em arXiv preprint arXiv:1904.07967 ICRA 2019 Workshop
  on Aerial Robotics}, 2019.

\bibitem{hulens2015choose}
D.~Hulens, T.~Goedem{\'e}, and J.~Verbeke, ``How to choose the best embedded
  processing platform for on-board uav image processing?,'' {\em Proceedings
  VISAPP 2015}, pp.~1--10, 2015.

\bibitem{bay2006surf}
H.~Bay, T.~Tuytelaars, and L.~Van~Gool, ``Surf: Speeded up robust features,''
  in {\em ECCV}, pp.~404--417, Springer, 2006.

\bibitem{lowe2004distinctive}
D.~G. Lowe, ``Distinctive image features from scale-invariant keypoints,'' {\em
  IJCV, Springer}, vol.~60, no.~2, pp.~91--110, 2004.

\bibitem{stumm2013probabilistic}
E.~Stumm, C.~Mei, and S.~Lacroix, ``Probabilistic place recognition with
  covisibility maps,'' in {\em IROS}, pp.~4158--4163, IEEE, 2013.

\bibitem{murillo2007surf}
A.~C. Murillo, J.~J. Guerrero, and C.~Sagues, ``Surf features for efficient
  robot localization with omnidirectional images,'' in {\em Proceedings of IEEE
  ICRA}, pp.~3901--3907, 2007.

\bibitem{oliva2006building}
A.~Oliva and A.~Torralba, ``Building the gist of a scene: The role of global
  image features in recognition,'' {\em Progress in brain research}, vol.~155,
  pp.~23--36, 2006.

\bibitem{murillo2009experiments}
A.~C. Murillo and J.~Kosecka, ``Experiments in place recognition using gist
  panoramas,'' in {\em ICCV Workshops}, pp.~2196--2203, IEEE, 2009.

\bibitem{singh2010visual}
G.~Singh and J.~Kosecka, ``Visual loop closing using gist descriptors in
  manhattan world,'' in {\em ICRA Omnidirectional Vision Workshop},
  pp.~4042--4047, 2010.

\bibitem{dalal2005histograms}
N.~Dalal and B.~Triggs, ``Histograms of oriented gradients for human
  detection,'' in {\em CVPR}, vol.~1, pp.~886--893, IEEE, 2005.

\bibitem{freeman1995orientation}
W.~T. Freeman and M.~Roth, ``Orientation histograms for hand gesture
  recognition,'' in {\em International workshop on automatic face and gesture
  recognition}, vol.~12, pp.~296--301, 1995.

\bibitem{mcmanus2014scene}
C.~McManus, B.~Upcroft, and P.~Newmann, ``Scene signatures: Localised and
  point-less features for localisation,'' {\em Robotics, Science and Systems
  Conference}, 2014.

\bibitem{zaffar2020cohog}
M.~Zaffar, S.~Ehsan, M.~Milford, and K.~McDonald-Maier, ``Cohog: A
  light-weight, compute-efficient, and training-free visual place recognition
  technique for changing environments,'' {\em IEEE Robotics and Automation
  Letters}, vol.~5, no.~2, pp.~1835--1842, 2020.

\bibitem{tolias2015particular}
G.~Tolias, R.~Sicre, and H.~J{\'e}gou, ``Particular object retrieval with
  integral max-pooling of cnn activations,'' {\em arXiv:1511.05879, ICLR},
  2016.

\bibitem{liu2016cross}
L.~Liu, C.~Shen, and A.~van~den Hengel, ``Cross-convolutional-layer pooling for
  image recognition,'' {\em IEEE transactions on pattern analysis and machine
  intelligence}, vol.~39, no.~11, pp.~2305--2313, 2016.

\bibitem{chen2014convolutional}
Z.~Chen, O.~Lam, A.~Jacobson, and M.~Milford, ``Convolutional neural
  network-based place recognition,'' {\em preprint arXiv:1411.1509}, 2014.

\bibitem{sermanet2014overfeat}
P.~Sermanet, D.~Eigen, X.~Zhang, M.~Mathieu, R.~Fergus, and Y.~LeCun,
  ``Overfeat: Integrated recognition, localization and detection using
  convolutional networks,'' in {\em 2nd International Conference on Learning
  Representations, ICLR 2014}, 2014.

\bibitem{chen2017deep}
Z.~Chen {\em et~al.}, ``Deep learning features at scale for visual place
  recognition,'' in {\em ICRA}, pp.~3223--3230, IEEE, 2017.

\bibitem{krizhevsky2012imagenet}
A.~Krizhevsky, I.~Sutskever, and G.~E. Hinton, ``Imagenet classification with
  deep convolutional neural networks,'' in {\em Advances in neural information
  processing systems}, pp.~1097--1105, 2012.

\bibitem{deng2009imagenet}
J.~Deng, W.~Dong, R.~Socher, L.-J. Li, K.~Li, and L.~Fei-Fei, ``Imagenet: A
  large-scale hierarchical image database,'' in {\em 2009 IEEE conference on
  computer vision and pattern recognition}, pp.~248--255, Ieee, 2009.

\bibitem{babenko2015aggregating}
A.~Babenko and V.~Lempitsky, ``Aggregating deep convolutional features for
  image retrieval,'' {\em arXiv preprint arXiv:1510.07493 ICCV}, 2015.

\bibitem{jaderberg2015spatial}
M.~Jaderberg, K.~Simonyan, A.~Zisserman, and K.~Kavukcuoglu, ``Spatial
  transformer networks,'' {\em arXiv preprint arXiv:1506.02025}, 2015.

\bibitem{arandjelovic2016netvlad}
R.~Arandjelovic, P.~Gronat, A.~Torii, T.~Pajdla, and J.~Sivic, ``Netvlad: Cnn
  architecture for weakly supervised place recognition,'' in {\em CVPR},
  pp.~5297--5307, 2016.

\bibitem{simonyan2014very}
K.~Simonyan and A.~Zisserman, ``Very deep convolutional networks for
  large-scale image recognition,'' {\em arXiv preprint arXiv:1409.1556}, 2014.

\bibitem{chen2018learning}
Z.~Chen, L.~Liu, I.~Sa, Z.~Ge, and M.~Chli, ``Learning context flexible
  attention model for long-term visual place recognition,'' {\em IEEE Robotics
  and Automation Letters}, vol.~3, no.~4, pp.~4015--4022, 2018.

\bibitem{khaliq2018holistic}
A.~Khaliq, S.~Ehsan, M.~Milford, and K.~McDonald-Maier, ``A holistic visual
  place recognition approach using lightweight cnns for severe viewpoint and
  appearance changes,'' {\em arXiv:1811.03032, IEEE T-RO}, 2018.

\bibitem{facil2019condition}
J.~M. Facil, D.~Olid, L.~Montesano, and J.~Civera, ``Condition-invariant
  multi-view place recognition,'' {\em arXiv preprint arXiv:1902.09516}, 2019.

\bibitem{hausler2019multi}
S.~Hausler, A.~Jacobson, and M.~Milford, ``Multi-process fusion: Visual place
  recognition using multiple image processing methods,'' {\em IEEE Robotics and
  Automation Letters}, vol.~4, no.~2, pp.~1924--1931, 2019.

\bibitem{canziani2016analysis}
A.~Canziani, A.~Paszke, and E.~Culurciello, ``An analysis of deep neural
  network models for practical applications,'' {\em arXiv preprint
  arXiv:1605.07678}, 2016.

\bibitem{huang2017speed}
J.~Huang, V.~Rathod, C.~Sun, M.~Zhu, A.~Korattikara, A.~Fathi, I.~Fischer,
  Z.~Wojna, Y.~Song, S.~Guadarrama, {\em et~al.}, ``Speed/accuracy trade-offs
  for modern convolutional object detectors,'' in {\em Proceedings of the IEEE
  conference on computer vision and pattern recognition}, pp.~7310--7311, 2017.

\bibitem{karki2019tango}
A.~Karki, C.~P. Keshava, S.~M. Shivakumar, J.~Skow, G.~M. Hegde, and H.~Jeon,
  ``Tango: A deep neural network benchmark suite for various accelerators,'' in
  {\em 2019 IEEE International Symposium on Performance Analysis of Systems and
  Software (ISPASS)}, pp.~137--138, IEEE, 2019.

\bibitem{palit2019uniform}
I.~Palit, Q.~Lou, R.~Perricone, M.~Niemier, and X.~S. Hu, ``A uniform modeling
  methodology for benchmarking dnn accelerators,'' in {\em 2019 IEEE/ACM
  International Conference on Computer-Aided Design (ICCAD)}, pp.~1--7, IEEE,
  2019.

\bibitem{fan2007power}
X.~Fan, W.-D. Weber, and L.~A. Barroso, ``Power provisioning for a
  warehouse-sized computer,'' {\em ACM SIGARCH computer architecture news},
  vol.~35, no.~2, pp.~13--23, 2007.

\bibitem{zhou2017places}
B.~Zhou, A.~Lapedriza, A.~Khosla, A.~Oliva, and A.~Torralba, ``Places: A 10
  million image database for scene recognition,'' {\em IEEE transactions on
  pattern analysis and machine intelligence}, vol.~40, no.~6, pp.~1452--1464,
  2017.

\bibitem{UPBoard}
``Up-cht01.''
  \url{https://www.aaeon.com/en/p/up-board-computer-board-for-professional-makers}.

\bibitem{RaspberryPi3}
``Raspberry pi 3 model b.''
  \url{https://www.raspberrypi.org/products/raspberry-pi-3-model-b/}.

\bibitem{OdroidXU4}
``Odroid xu4.'' \url{https://www.odroid.co.uk/odroid-xu4}.

\bibitem{phdthesis}
A.~Abdelbaki, M.~Bennewitz, and R.~Sabverzavi, {\em ConvNet Features for
  Lifelong Place Recognition and Pose Estimation in Visual SLAM}.
\newblock PhD thesis, 03 2018.

\bibitem{merrill2018lightweight}
N.~Merrill and G.~Huang, ``Lightweight unsupervised deep loop closure,'' {\em
  arXiv preprint arXiv:1805.07703, Robotics Science and Systems Conference},
  2018.

\bibitem{zaffar2018memorablemaps}
M.~Zaffar, S.~Ehsan, M.~Milford, and K.~McDonald-Maier, ``Memorable maps: A
  framework for re-defining places in visual place recognition,'' {\em
  https://arxiv.org/abs/1811.03529}, 2018.

\bibitem{siepert_wellish_2019}
B.~Siepert and I.~Wellish, ``Adafruit ina260 current voltage power sensor
  breakout,'' May 2019.

\bibitem{porav2018adversarial}
H.~Porav, W.~Maddern, and P.~Newman, ``Adversarial training for adverse
  conditions: Robust metric localisation using appearance transfer,'' in {\em
  2018 IEEE International Conference on Robotics and Automation (ICRA)},
  pp.~1011--1018, IEEE, 2018.

\end{thebibliography}
}

\end{document}